\pdfoutput=1

\documentclass[11pt]{article}

\usepackage{acl}

\usepackage{times}
\usepackage{latexsym}

\usepackage[T1]{fontenc}

\usepackage[utf8]{inputenc}

\usepackage{microtype}
\usepackage{booktabs}
\usepackage{amsmath}
\usepackage{graphicx}
\usepackage{caption}
\usepackage{threeparttable}
\usepackage{tablefootnote}
\usepackage{footnote}

\usepackage{dsfont}

\title{Selecting Stickers in Open-Domain Dialogue through Multitask Learning}

\author{Zhexin Zhang$^{12}$\Thanks{~Work done during internship at WeChat AI.}~, Yeshuang Zhu$^2$, Zhengcong Fei$^2$, Jinchao Zhang$^2$, Jie Zhou$^2$ \\
\small{$^1$The CoAI group, DCST; $^1$Institute for Artificial Intelligence; $^1$State Key Lab of Intelligent Technology and Systems;}\\
\small{$^1$Beijing National Research Center for Information Science and Technology;} 
\small{$^1$Tsinghua University, Beijing 100084, China.}\\
\small{$^2$Pattern Recognition Center, WeChat AI, Tencent Inc, China.}\\
\small{\texttt{zx-zhang18@mails.tsinghua.edu.cn,}}
\small{\texttt{\{yshzhu,dayerzhang,withtomzhou\}@tencent.com,}}\\
\small{\texttt{feizhengcong@ict.ac.cn,}}}

\begin{document}
\maketitle
\begin{abstract}
With the increasing popularity of online chatting, stickers are becoming important in our online communication. Selecting appropriate stickers in open-domain dialogue requires a comprehensive understanding of both dialogues and stickers, as well as the relationship between the two types of modalities. To tackle these challenges, we propose a multitask learning method comprised of three auxiliary tasks to enhance the understanding of dialogue history, emotion and semantic meaning of stickers. Extensive experiments conducted on a recent challenging dataset show that our model can better combine the multimodal information and achieve significantly higher accuracy over strong baselines. Ablation study further verifies the effectiveness of each auxiliary task. Our code is available at \url{https://github.com/nonstopfor/Sticker-Selection}.

\end{abstract}

\section{Introduction}
With the development of mobile messaging apps (e.g., WhatsApp and Messenger), visual content is getting more and more frequently used in our daily conversation, such as emojis and stickers. Compared with emojis, stickers are larger images consisting of drawing characters, symbolic icons, and text titles, and are hence more expressive and versatile \cite{konrad2020sticker}. Users send stickers along with text to show intimacy, express strong emotion, and 
experience the enjoyment of creativity \cite{tang2019emoticon}.

\begin{figure}[!t]
\includegraphics[width=0.9\linewidth]{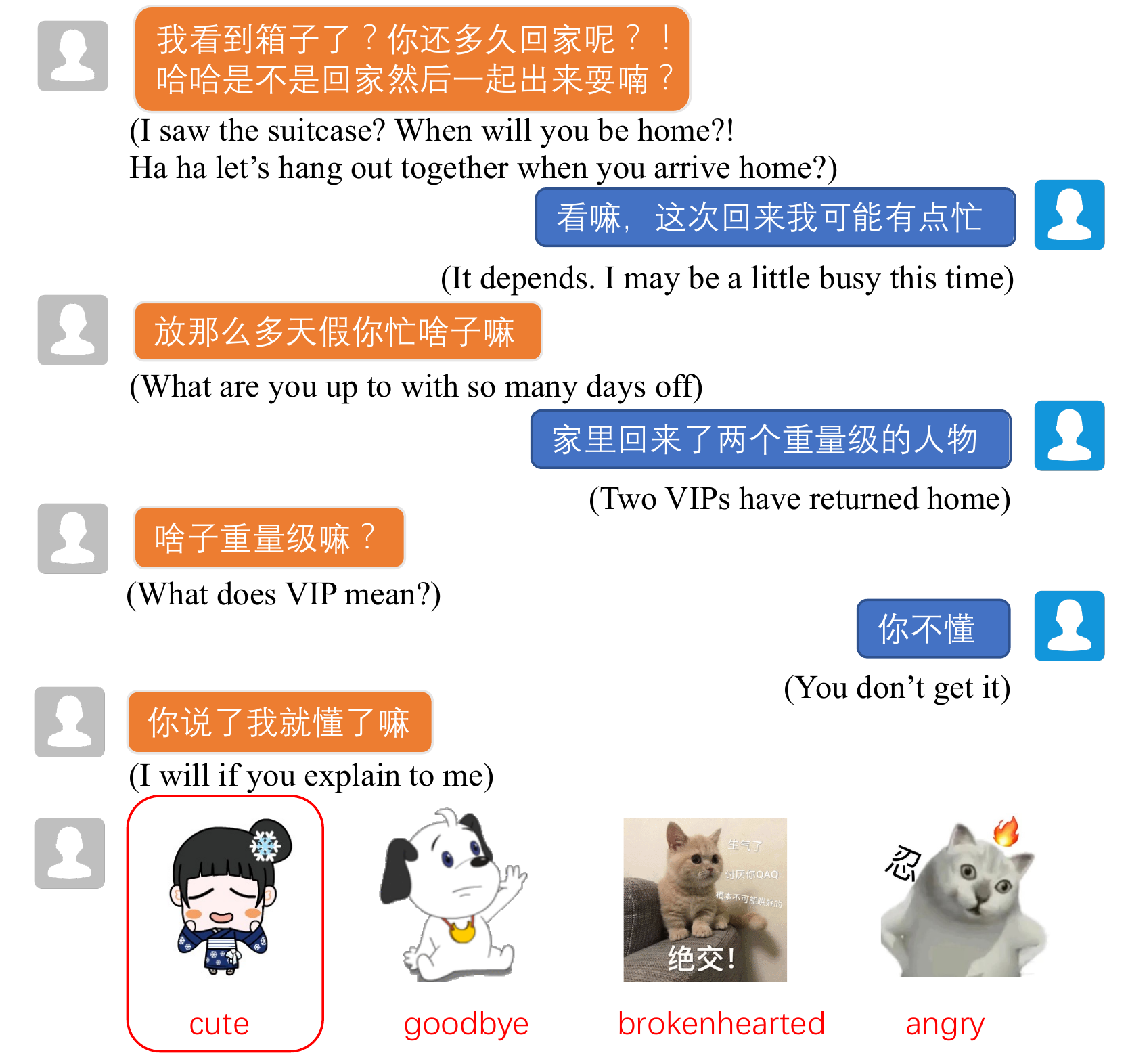}
  \caption{An example of the sticker selection task. Given a dialogue history, the model needs to add a sticker to the last textual message which is the most appropriate one among a collection of candidate stickers (the one marked in the red rectangle). The words below in red denote the emotion or meaning of each sticker.}
  \label{fig:goodcase1}
\end{figure}

Despite the importance of stickers in daily communication, selecting stickers in open-domain dialogue hasn't been widely explored. In this paper, we address the task of selecting an appropriate sticker from a candidate set for an open-domain multi-turn dialogue. This task is a typical setting for various applications, e.g., automatically recommending stickers in messaging apps and building more interesting and human-like chatbots which could respond with stickers. As shown in Figure~\ref{fig:goodcase1}, this task requires an understanding of dialogue context, emotion and semantic meaning of stickers, and a jointly modeling ability for the multimodal information. Only a few previous works have explored this task \cite{gao2020learning,wang2021animated}. However, existing models are only trained on an end-to-end matching objective and lacks finer-grained supervision signals which could guide models to understand multimodal information better.



Considering the challenges of this task and the shortcomings of previous work, we propose a novel multitask learning method to improve sticker selection in open-domain multi-turn dialogue. We design three auxiliary tasks: \textbf{1) masked context prediction}, which uses multimodal context to predict masked tokens in the dialogue history, aiming to understand the dialogue in the presence of the sticker; \textbf{2) sticker emotion classification}, which utilizes the sticker's contextualized representation to predict its emotion, aiming to improve the model's understanding of sticker emotion; \textbf{3) sticker semantic prediction}, which explicitly instills semantic understanding of stickers by training the model to reconstruct a sticker's semantic label based on the multimodal inputs. Moreover, all these tasks help improve our model's joint modeling capability, as both our model architecture and task design require multimodal inputs and deep interactions between them. We evaluate the performance of our method on a recently proposed and challenging dataset. Extensive experiments show that our multitask method achieves state-of-the-art performance.

There are two contributions of this paper:
\begin{itemize}
	\item We propose a multitask learning method to help select appropriate stickers in open-domain multi-turn dialogue.
	\item Experiment results on a challenging dataset demonstrate the effectiveness of each auxiliary task and combining all the tasks achieves state-of-the-art performance.
\end{itemize}

\section{Related Work}
\paragraph{Sticker selection.} Previous works proposed to recommend emojis in dialogue systems based on textual or multimodal context \cite{barbieri2018multimodal,xie2016neural,barbieri2017emojis}. However, emojis are limited in variety and are much less expressive than stickers.
\citet{laddha2020understanding} retrieved stickers for generated text utterances by simply matching the text tags of stickers. Several works have proposed improved matching methods for stickers.
\citet{gao2020learning} utilized co-attention to capture the interaction between a sticker and each utterance, and used a fusion network to combine the features. 
\citet{wang2021animated} followed the matching framework of CLIP \cite{radford2021learning} and designed a multimodal encoder for animated GIFs.
\citet{fei2021towards} proposed to generate special sticker tokens along with text utterances using one single GPT \cite{wang2020large} for emotion prediction and retrieval of stickers.
However, existing models are only trained on an end-to-end matching objective that implicitly guides the models to understand multimodal information.
In our work, we design finer-grained auxiliary tasks that instill knowledge of stickers and their contextualized usage in a more efficient way.

\paragraph{Visual Dialogue.}
Visual dialogue is a task to answer questions about the factual content of the real-world image \cite{liang-etal-2021-maria,das2017visual,das2017learning}. In contrast, selecting appropriate stickers in open-domain dialogue requires understanding sentiment and semantic expression of user-generated, artistic style images.


\section{Method}
\label{method}

\begin{figure}[!t]
\includegraphics[width=\linewidth]{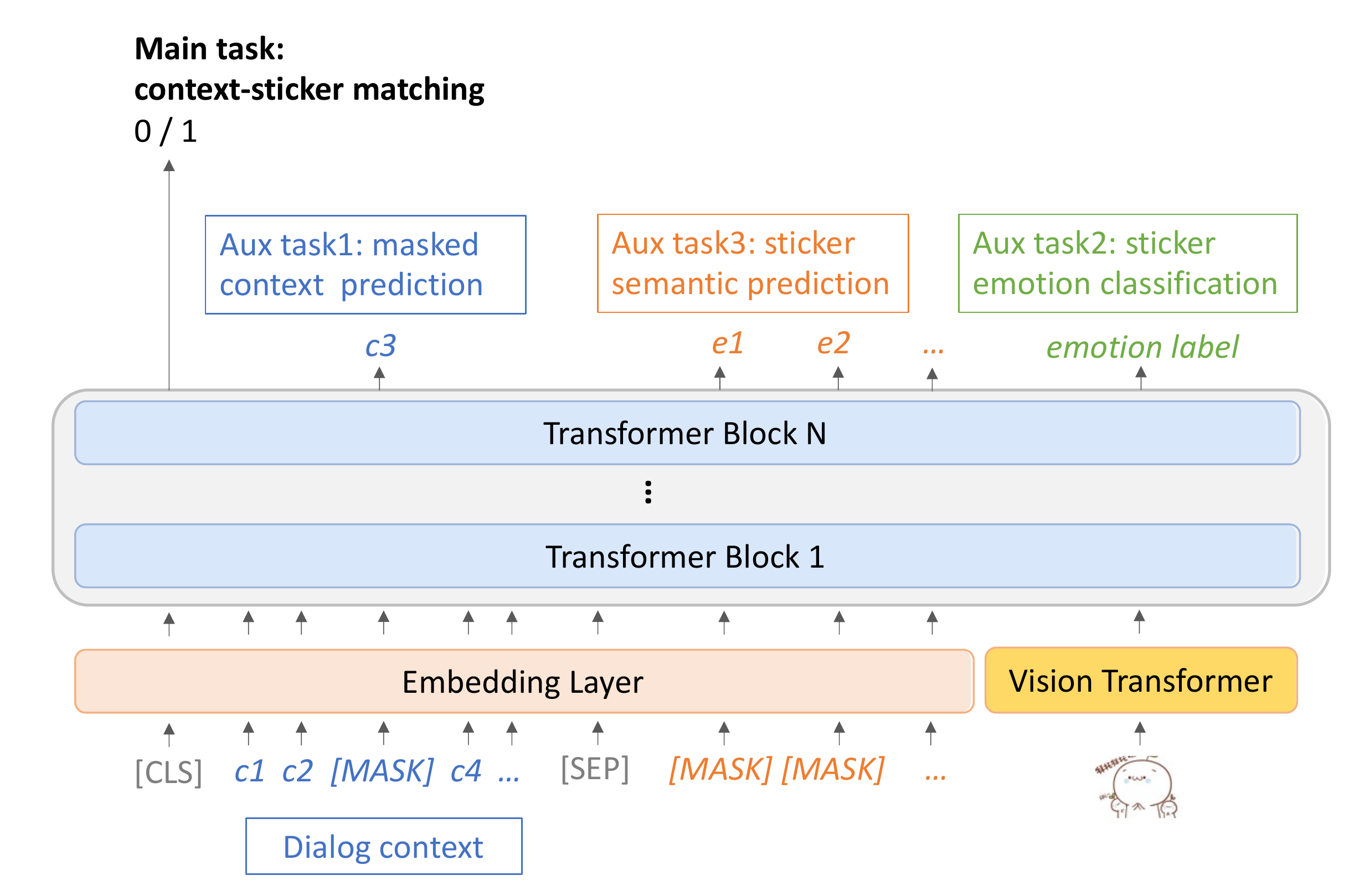}
  \caption{An overview of our training task design. The base model architecture is a multimodal BERT that learns to predict whether the candidate sticker is appropriate given the dialogue context. Three auxiliary tasks are proposed to enhance the model's ability to understand multimodal input. $c_i$ and $e_i$ represent tokens of dialogue context and semantic label respectively.}
  \label{fig:model}
\end{figure}


\subsection{Task Definition}
We assume there is a multi-turn dialogue context $C=\{u_1,...,u_N\}$, and a candidate sticker set $S=\{s_1,...,s_M\}$, where $u_i$ represents the $i$-th utterance in the dialogue, and $s_i$ represents the $i$-th candidate sticker. $N$ is the number of utterances in the dialogue and $M$ is the number of candidate stickers. In this work, we suppose that there is only one appropriate sticker $s^* \in S$, and $s^*$ and $u_N$ belong to the same speaker. The goal is to train a model that can select the right sticker $s^*$ among all candidates $S$ given the dialogue history $C$.

\subsection{Method Overview}
An overview of the design of our training tasks is shown in Figure \ref{fig:model}. Our main task is to decide whether the candidate sticker is appropriate given the dialogue context. To accomplish this task, we concatenate the embedded dialogue context and the sticker embedding as inputs to BERT. Then we apply a binary classification layer on top of the hidden state of the [CLS] token. In order to enhance the model's ability to understand the multimodal input, we design three auxiliary tasks: \textbf{1) masked context prediction}, which improves the model's understanding of dialogue context; \textbf{2) sticker emotion classification}, which aims to make the model better understand sticker's emotion; \textbf{3) sticker semantic prediction}, which instills semantic information of stickers to the model. Next, we will introduce our three auxiliary tasks in detail.



\subsection{Task 1: Masked Context Prediction}
The masked context prediction task follows the masked language modeling (MLM) task in BERT \cite{devlin-etal-2019-bert}. One difference is that we additionally append the embedding of the appropriate sticker to the input embeddings. In this way, the model can learn to utilize stickers for dialogue reconstruction, and thus the interaction between the two modalities is enhanced. The loss for this task is denoted as $\mathcal{L}_{ctx}$, and takes the same form of cross-entropy loss as in the original MLM task.

\subsection{Task 2: Sticker Emotion Classification}
In the dataset we used, stickers are annotated with one context-dependent emotion, which means one sticker could have different emotions in different dialogue contexts. Therefore, we design a sticker emotion classification task to enable the model to utilize the text and sticker information simultaneously for understanding sticker emotion. Specifically, we take the hidden state corresponding to the sticker and apply a softmax layer with cross-entropy loss on top of it for emotion classification. The loss for this task is denoted as $\mathcal{L}_{emo}$.

\subsection{Task 3: Sticker Semantic Prediction}
Task 1 and Task 2 emphasize learning the implicit meaning of stickers and their correlation with dialogue text. However, many stickers express a clear intention that indicates their proper usage context, e.g., greetings and declines. We believe empowering our model to predict and utilize the semantic meaning of stickers is beneficial for our task. Hence, we further design a semantic label prediction task. We modify our model's inputs by inserting a fixed-length sequence of [MASK] tokens after the dialogue. The model is trained to recover the label text from the hidden states of the [MASK] tokens. The loss is formulated as the sum of cross-entropy loss for each token in label and is denoted as $\mathcal{L}_{sem}$.
Since the dataset we used has no ground truth semantic labels for stickers, we take the textual information recognized by an OCR tool as semantic labels for stickers.
Note that $\mathcal{L}_{sem}$ is only applied for those stickers with text recognized.

\subsection{Total Loss}
Besides the above three auxiliary tasks, our main task is a binary classification of whether a candidate sticker is appropriate given the dialogue context. We take all dialogue-sticker pairs in the dataset as positive samples and randomly sample stickers to create an equal number of negative samples. The cross-entropy loss is denoted as $\mathcal{L}_{main}$.

Our final loss is a combination of the four loss:

\begin{align}
    \mathcal{L}=\mathcal{L}_{main}+\alpha\mathcal{L}_{ctx}+\beta\mathcal{L}_{emo}+\gamma\mathcal{L}_{sem}
\end{align}
where $\alpha,\beta,\gamma$ are manually tuned hyperparameters.
\section{Experiments}

\subsection{Dataset}

We use the Chinese version of the MOD dataset from DSTC10-Track1\footnote{\url{https://openai.weixin.qq.com/dstc/DescriptionEN}. See Appendix~\ref{sec:datastats} for more details.}. The dataset is grounded in a dialogue scenario and contains various stickers with contextualized emotion annotation. We split each dialogue into several samples, each containing a text sequence of dialogue history and an accompanying sticker. Note that this dataset is revised from the unpublished one used in \citet{fei2021towards}.

\begin{savenotes}
\begin{table*}[!ht] 
\small
	\begin{center}	\setlength{\tabcolsep}{1.5mm}{
		\begin{tabular}{l|llll|llll}
			\toprule
			&\textbf{R$_{10}$@1}& \textbf{R$_{10}$@2}&\textbf{R$_{10}$@5}&\textbf{MRR$_{10}$}&\textbf{R$_\text{ALL}$@1}& \textbf{R$_\text{ALL}$@2}&\textbf{R$_\text{ALL}$@5}&\textbf{MRR$_\text{ALL}$}\\
		
			\midrule
		
				\multicolumn{5}{l}{\emph{easy test}}\\ \midrule 
			SRS\footnote{We only provide the results with 10 candidate stickers as their public code does.}&30.51&54.24&71.28&48.15&-&-&-&-\\
			MOD-GPT&31.20&54.81&72.13&49.20&5.10&9.05&15.57&11.46\\
			CLIP&38.44&56.45&82.27&56.76&6.00&9.39&16.61&12.69\\
			MMBERT&45.44&66.78&90.95&64.03&5.69&10.08&19.44&13.98\\
			MMBERT+\textit{ctx}&47.06&67.34&90.76&65.00&5.91&10.26&20.72&14.44\\
			MMBERT+\textit{ctx}+\textit{emo}&48.80&\textbf{70.67}&\textbf{92.29}&\textbf{66.88}&6.07&11.26&22.02&15.22\\
			MMBERT+\textit{ctx}+\textit{emo}+\textit{sem}&\textbf{49.14$^{**}$}&69.46$^{**}$&91.76$^{**}$&66.67$^{**}$&\textbf{7.40$^{**}$}&\textbf{12.07$^{**}$}&\textbf{22.08$^{**}$}&\textbf{15.99$^{**}$}\\
			
			\midrule
					\multicolumn{5}{l}{\emph{hard test}}\\ \midrule 
					
			SRS&23.85&45.30& 63.52&40.33&-&-&-&-\\
			MOD-GPT&25.50&49.22&64.03&40.51&3.52&6.12&12.76&9.23\\
			CLIP&32.81&48.55&76.17&51.28&5.79&8.91&15.14&11.55\\
			MMBERT&32.47&50.40&78.32&51.88&3.90&6.62&13.15&9.71\\
			MMBERT+\textit{ctx}&33.11&51.04&78.98&52.60&4.21&7.71&13.82&10.38\\
			MMBERT+\textit{ctx}+\textit{emo}&35.39&52.26&78.14&53.65&4.87&8.18&14.66&11.06\\
			MMBERT+\textit{ctx}+\textit{emo}+\textit{sem}&\textbf{36.64$^{**}$}&\textbf{55.48$^{**}$}&\textbf{80.78$^{**}$}&\textbf{55.40$^{**}$}&\textbf{6.06}&\textbf{9.65$^{*}$}&\textbf{15.79}&\textbf{12.40$^{**}$}\\
			\bottomrule
		\end{tabular} }
	\end{center}
	{\captionsetup{font=small} \caption{Performance of the models on DSTC10 dataset. All the numbers are scaled by 100. The easy test set contains only the same stickers seen during training, while the hard test set has unseen stickers. The footnotes $_{10}$ and $_\text{ALL}$ indicate the numbers of candidate stickers considered for each train and test case, which are 10 (ground truth sticker plus 9 randomly sampled stickers) or all available stickers respectively. R@k is the recall rate of top-k predicted stickers and MRR the Mean Reciprocal Rank of ground truth stickers. The abbreviations \textit{ctx}, \textit{emo} and \textit{sem} correspond to the auxiliary task 1, 2 and 3 respectively in Section~\ref{method}. A paired t-test is conducted between the full model (MMBERT+\textit{ctx}+\textit{emo}+\textit{sem}) and CLIP ($^{*}$: $p < 0.05$, $^{**}$: $p < 0.01$).}
		\label{tab:result}}
\end{table*}
\end{savenotes}

\subsection{Baselines}

We compare our model with the following baselines from recent related work: \textbf{1) SRS} \cite{gao2020learning}, which encodes dialogue history and candidate sticker separately, and then employs a deep interaction network and a fusion network to score each candidate sticker; \textbf{2) MOD-GPT} \cite{fei2021towards}, which uses one single GPT to generate response text and match sticker; \textbf{3) CLIP}, which finetunes pretrained CLIP \cite{radford2021learning} for sticker selection using the same contrastive loss.

\begin{figure}[!ht]
\includegraphics[width=\linewidth]{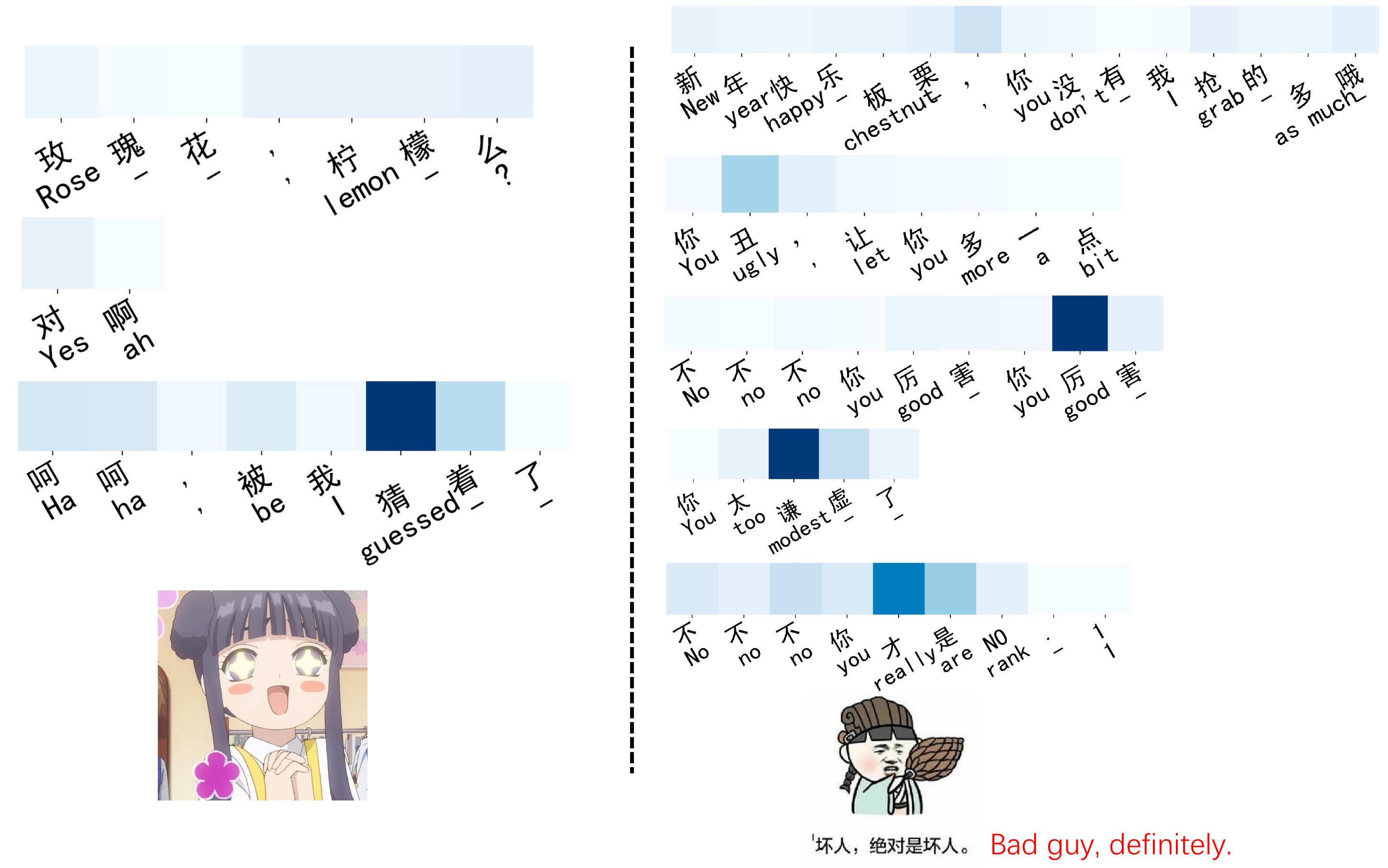}
  \caption{Examples of word saliency in the dialogue history. Word saliency is computed as Frobenius norm of its gradient with regard to the main task loss. Darker color indicates the word is more important. The words in red denote the text in the sticker.}
  \label{fig:grad}
\end{figure}

\begin{figure}[!ht]
\includegraphics[width=\linewidth]{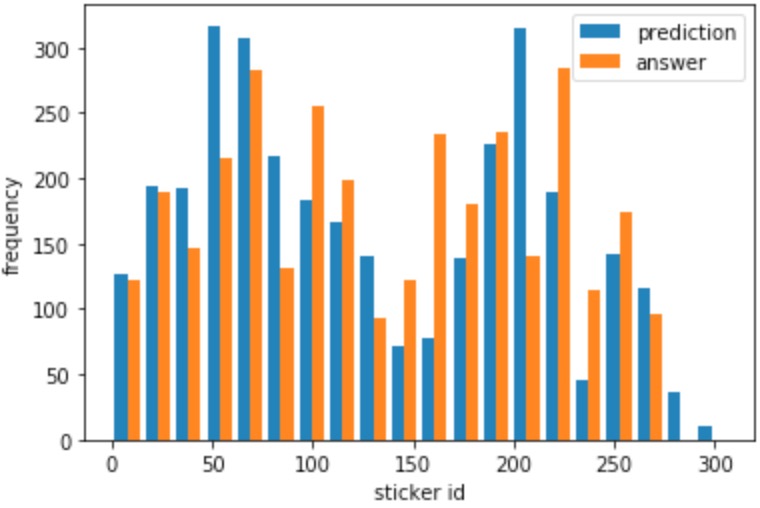}
  \caption{The diversity of the predicted and ground truth stickers in the easy test set.}
  \label{fig:pred_diversity}
\end{figure}

\begin{figure}[!ht]
\includegraphics[width=\linewidth]{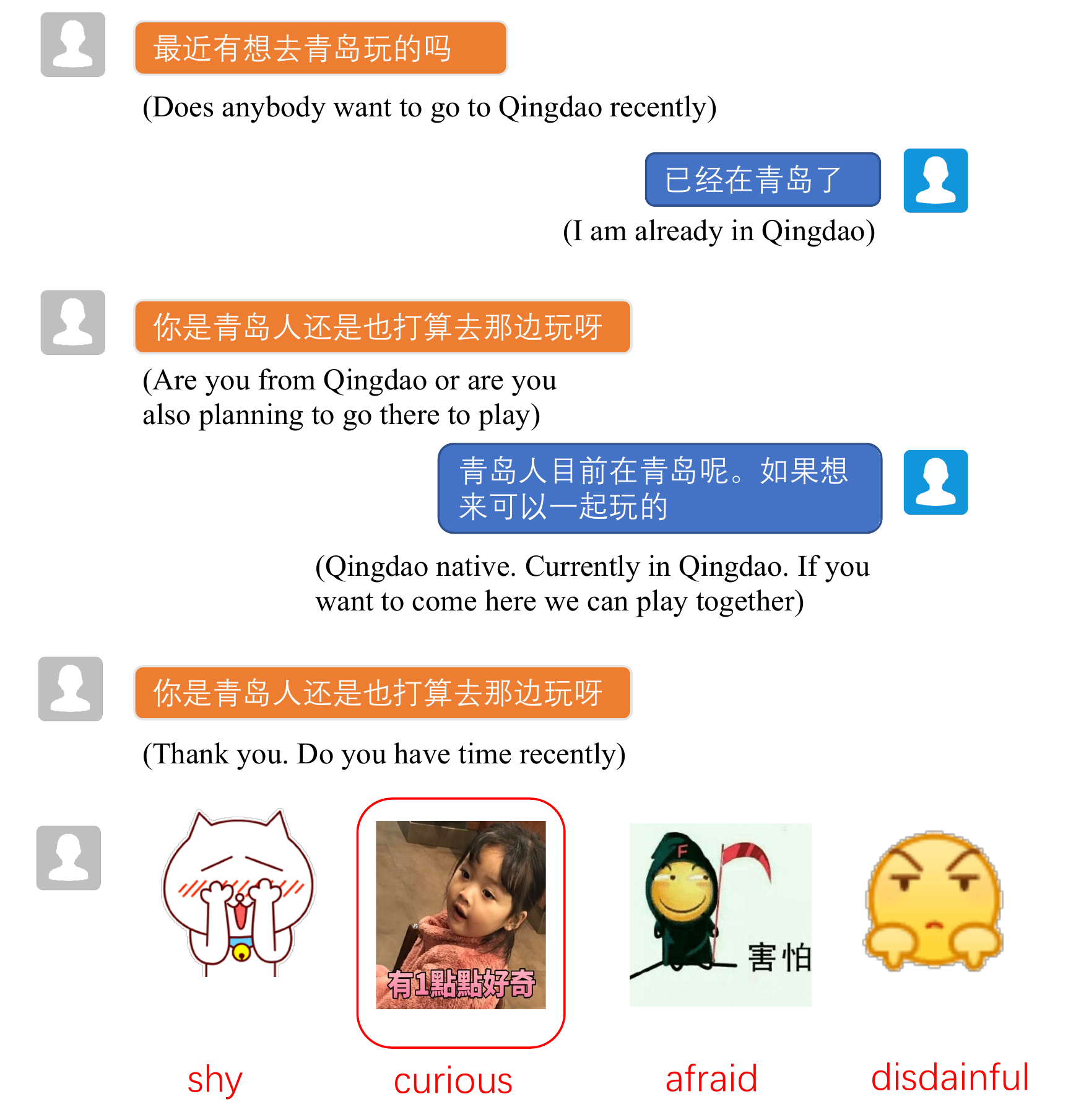}
  \caption{A failing case of our model. The leftmost sticker is selected by our model among the four candidate stickers. The appropriate sticker is marked with a red rectangle. The red words explain the stickers' emotions or meanings.}
  \label{fig:badcase1}
\end{figure}

\begin{figure}[!ht]
\includegraphics[width=\linewidth]{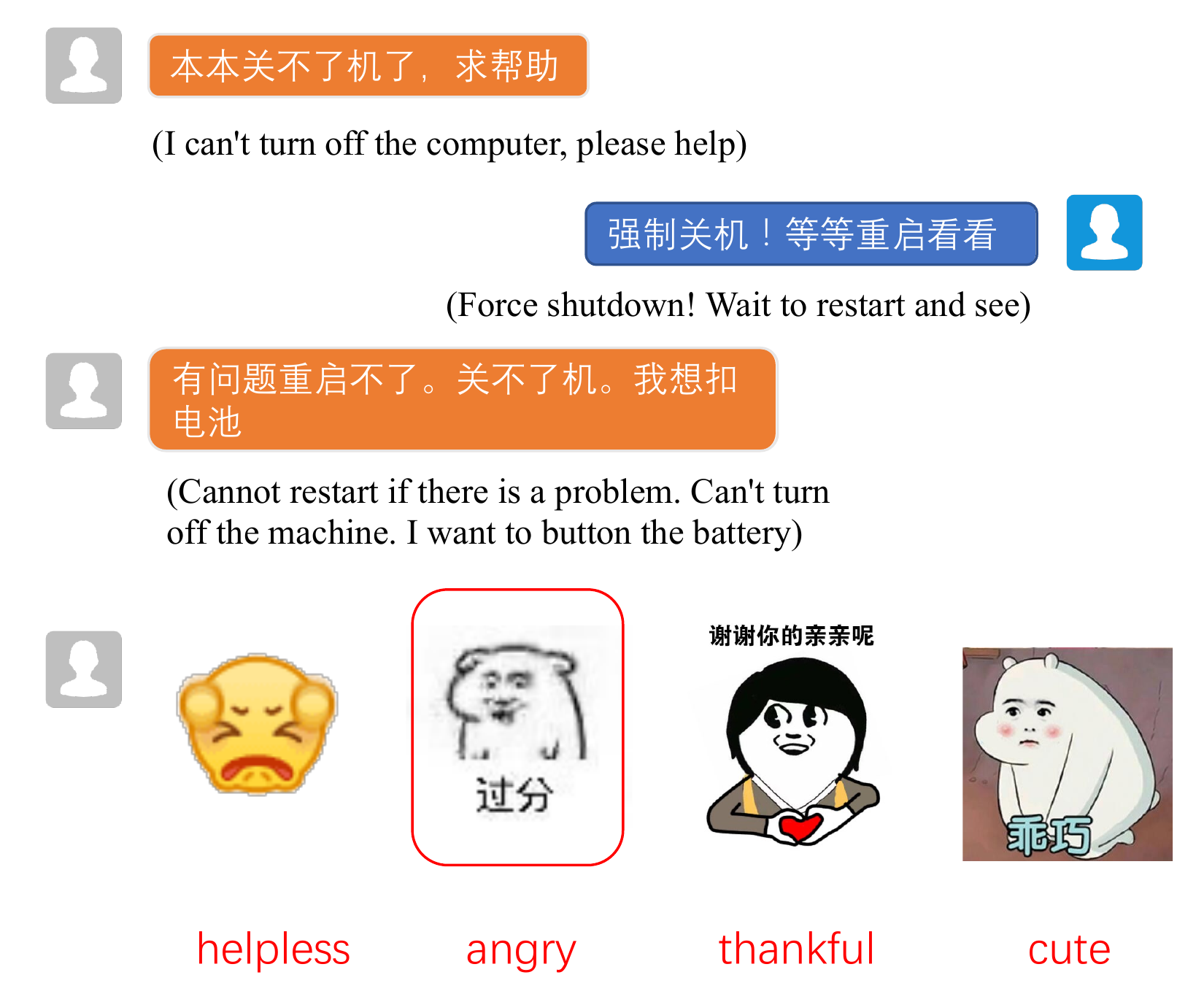}
  \caption{A case in which our model's prediction is not the same as the answer but also appropriate. The leftmost sticker is selected by our model among the four candidate stickers. The appropriate sticker is marked with a red rectangle. The red words explain the stickers' emotions or meanings.}
  \label{fig:case2}
\end{figure}

\subsection{Results and Analysis}
The result is shown in Table~\ref{tab:result}. Our full model (MMBERT+\textit{ctx}+\textit{emo}+\textit{sem}) outperforms all baselines on two test sets, and achieves the best performance in almost all settings. As expected, all the results get worse on the hard test set and when selecting one amongst all stickers. As only one out of the numerous and various online stickers is considered correct, the task is inherently challenging.
We find that CLIP is a strong baseline due to its better generalization ability on the hard set, compared with our base model which has no auxiliary task (MMBERT). This may be because CLIP is pretrained on a large number of image-text pairs. However, with multitask learning, our full model outperforms CLIP, although BERT has never seen images during pretraining. Thus, we conclude that our multitask learning method can improve sticker selection by explicitly guiding the model to understand multimodal information.

We also perform an ablation study to verify the effect of each auxiliary task. 
A clear trend emerges that the performance improves as each auxiliary training task is added to MMBERT, verifying the efficacy of our task design.
One exception is that MMBERT+\textit{ctx}+\textit{emo} performs slightly better than our full model in terms of R$_{10}$@2, R$_{10}$@5, and MRR$_{10}$. However, the inconsistency disappears when considering all stickers as candidates. Furthermore, our full model performs significantly better on the hard test set which contains unseen stickers. Hence, we conclude that introducing semantic information improves the model's generalization ability.
We also find that our full model achieves 60\% accuracy on the validation set for the auxiliary sticker emotion classification task with 52 emotion labels in total, which is reasonable and confirms our model can learn from the auxiliary tasks.

We visualize the saliency of different words in the dialogue history in Figure \ref{fig:grad}, which shows that the more relevant words (e.g., \textit{guessed}, \textit{good} and \textit{modest}) in the dialogue history contribute more to our model's prediction. Notably, our model could attend to some distant words (e.g., \textit{good}), not just the words inside the previous utterance. 

We also analyze the prediction diversity of our full model. As shown in Figure \ref{fig:pred_diversity}, the predictions of our model are diverse in general, covering almost all stickers in the whole candidate set. We note that a few stickers are predicted significantly more times than other stickers, which is because they appear much more frequently than other stickers in the training set. We leave addressing the imbalance problem of the training set as our future work.




\subsection{Case Study}
We present a successful case in Figure~\ref{fig:goodcase1}, where the ground truth sticker has no OCR information, making it challenging for the model to understand its semantic meaning. Moreover, the model needs to understand that the dialogue is in a delighted context, and the stickers' emotions and meanings in order to distinguish the most appropriate sticker from the others. This case suggests our model has a good understanding of dialogue history and sticker emotion and semantic meaning with the help of auxiliary tasks.

We show a failing case of our model in Figure \ref{fig:badcase1}. In this case, the appropriate sticker never appears in the training set. Considering the hard test set is more challenging than the easy test set, improving the generalization ability of our model is thus an important direction of future work. The same is true for baselines.

In the dataset we used, only one sticker is considered correct. However, we observe cases where the model's selection is not the same as the answer but is also appropriate. An example is shown in Figure \ref{fig:case2}. Therefore, the results in Table  \ref{tab:result} indicate a lower bound performance and our model may perform better in practice. 

\section{Conclusion}
In this paper, we address the challenging task of selecting appropriate stickers in open-domain multi-turn dialogue. We propose a multitask learning method with three auxiliary tasks to enhance the understanding of dialogues and stickers. Experiments show that our model outperforms strong baselines, confirming the effectiveness of our multitask learning method for sticker selection. Although our experiments are conducted on a Chinese dataset, our methods are expected to work for other languages.

\bibliography{anthology,custom}
\bibliographystyle{acl_natbib}

\appendix
\label{sec:appendix}

\begin{table*}[t] 
\small
	\begin{center}	\setlength{\tabcolsep}{1.5mm}{
		\begin{tabular}{l|cccc|cccc}
			\toprule
			&\textbf{R$_{10}$@1}& \textbf{R$_{10}$@2}&\textbf{R$_{10}$@5}&\textbf{MRR$_{10}$}&\textbf{R$_\text{ALL}$@1}& \textbf{R$_\text{ALL}$@2}&\textbf{R$_\text{ALL}$@5}&\textbf{MRR$_\text{ALL}$}\\
			\midrule
		
				\multicolumn{5}{l}{\emph{easy test}}\\ \midrule 
			MMBERT+\textit{ctx}+\textit{emo}&48.80&70.67&92.29&66.88&6.07&11.26&22.02&15.22\\
			MMBERT+\textit{ctx}+\textit{emo}+\textit{sem}&49.14&69.46&91.76&66.67&7.40&12.07&22.08&15.99\\
			MMBERT+\textit{ctx}+\textit{emo}+\textit{sem}-OCR&49.95&70.89&92.13&67.39&6.63&12.07&22.18&15.80\\
			MMBERT+\textit{ctx}+\textit{emo}+\textit{sem}-OCR+data&47.06&67.50&91.07&65.12&6.03&9.74&19.75&14.23\\
			
			\midrule
					\multicolumn{5}{l}{\emph{hard test}}\\ \midrule 
			MMBERT+\textit{ctx}+\textit{emo}&35.39&52.26&78.14&53.65&4.87&8.18&14.66&11.06\\
			MMBERT+\textit{ctx}+\textit{emo}+\textit{sem}&36.64&55.48&80.78&55.40&6.06&9.65&15.79&12.40\\
			MMBERT+\textit{ctx}+\textit{emo}+\textit{sem}-OCR&32.87&50.03&76.07&51.51&4.58&7.74&13.70&10.44\\
			MMBERT+\textit{ctx}+\textit{emo}+\textit{sem}-OCR+data&33.42&50.94&78.78&52.49&4.75&7.80&14.31&10.75\\

			\bottomrule
		\end{tabular} }
	\end{center}
	{\caption{Effect of incorporating semantic label prediction and OCR feature on DSTC10 dataset. All the numbers are scaled by 100. The easy test set only contains stickers ever seen in the training set, while the hard test set contains stickers unseen during training. R$_{10}$@k and R$_\text{ALL}$@k mean recall rate of ground truth stickers from top-k stickers chosen by the models, given a candidate set of 10 or all available stickers respectively. MRR$_{10}$ and MRR$_\text{ALL}$ represent Mean Reciprocal Rank of ground truth stickers among 10 or all available stickers. MMBERT+\textit{ctx}+\textit{emo}+\textit{sem}-OCR means not using OCR information for other tasks except sticker semantic prediction. MMBERT+\textit{ctx}+\textit{emo}+\textit{sem}-OCR+data means not using OCR information for other tasks except sticker semantic prediction and adds extra sticker-description pairs for sticker semantic prediction task. }
		\label{tab:result2}}
\end{table*}

\begin{table}[]
\footnotesize
\centering
\begin{tabular}{@{}c|cccc@{}}
\toprule
                                  & Train & Valid & Easy test & Hard test \\ \midrule
\# samples                        & 211575      &  3542     &   3215        & 7028          \\
\# emo samples & 209890 & 3495 & - & - \\
\# utterances                     & 1666208      &  26040     & 25447          &    59773       \\
\# tokens                         &  10400     &    2718   &    2780       &    3818       \\
\# stickers                       &  283     & 249      & 239          & 278           \\
Avg. \# utterances  &  7.88     & 7.35      &     7.92      &     8.50      \\
Avg. \# tokens &  18.42     &    12.47   &  12.91         &  14.54         \\ \bottomrule
\end{tabular}
\caption{Dataset statistics. Easy test set's stickers all appear in the train set, while the hard test set contains stickers which don't appear in the train set. One original dialogue could be split into several samples, each containing one sticker response. The token num is computed by the tokenizer of BERT. \# emo samples means the number of samples containing emotion annotation. Avg. \# tokens means the average number of tokens for each utterance.}
\label{tab:stat}
\end{table}

\section{Dataset Details}
\label{sec:datastats}
Statistics of the dataset are shown in Table \ref{tab:stat}. There are 307 stickers in total and 228 out of them have textual information extracted by OCR. For stickers without emotion labels or semantic labels, we simply ignore the emotion classification loss or the semantic prediction loss. A better way to deal with the missing labels is left as future work.

For each dialogue sample, we ignore stickers in the middle of the dialogue history, as we found in preliminary experiments that removing them has no significant impact on the performance.

\section{Implementation Details}
For all the models implemented by ourselves in our experiments, we set the batch size to 8 and use AdamW optimizer with cosine scheduler.
For the CLIP baseline, as there is no available CLIP model especially pretrained in Chinese, we use a multilingual version adapted via knowledge distillation \cite{reimers2020making}\footnote{\url{https://www.sbert.net/docs/pretrained_models.html\#image-text-models}}. We cut the dialogue history to take only the last sentence in order to fit the length limit of CLIP's text encoder. We also tried to use the last two or more sentences, but found that the performance decreased.
All BERT-based models and MOD-GPT use the image encoder in the  CLIP baseline.
We set the CLIP image encoder's learning rate to 5e-7 and the text encoder's learning rate to 9e-6.
For BERT-based models, we set the learning rate to 9e-6 and fix the image encoder following  \cite{fei2021towards}.  The maximum epoch is set to 10. For the total loss, $\alpha$ is set to 0.05, $\beta$ is set to 0.2 and $\gamma$ is set to 0.1. All the hyperparameters are selected based on the validation set. The maximum training time for one epoch is about 5 hours on one single V100 GPU.

\begin{table*}[!t] 

	\begin{center}	\setlength{\tabcolsep}{1.5mm}{
		\begin{tabular}{l|llll}
			\toprule
			&\textbf{R$_\text{ALL}$@1}& \textbf{R$_\text{ALL}$@2}&\textbf{R$_\text{ALL}$@5}&\textbf{MRR$_\text{ALL}$}\\
		
			\midrule
		
				\multicolumn{5}{l}{\emph{easy test}}\\ \midrule 
			
			CLIP&7.49/2.95/\textbf{4.54}&11.23/6.09/\textbf{5.14}&18.21/13.23/\textbf{4.98}&14.29/9.54/\textbf{4.75}\\
			
			MMBERT+\textit{ctx}+\textit{emo}+\textit{sem}&8.69/4.85/3.84&12.80/10.75/2.05&19.69/27.12/-7.43&15.89/16.28/-0.39\\
			
			\midrule
					\multicolumn{5}{l}{\emph{hard test}}\\ \midrule 

			CLIP&6.82/3.58/\textbf{3.24}&10.23/6.46/\textbf{3.77}&16.60/12.66/\textbf{3.94}&12.74/9.26/\textbf{3.48}\\
			
			MMBERT+\textit{ctx}+\textit{emo}+\textit{sem}&6.95/4.50/2.45&10.68/7.77/2.91&16.48/14.62/1.86&13.24/10.91/2.33\\
			\bottomrule
		\end{tabular} }
	\end{center}
	{ \caption{Performance of our full model and CLIP on the divided dataset. \textbf{The three values divided by / correspond to the performance on the sub dataset where stickers have text, the performance on the sub dataset where stickers don't have text and their difference}.  All the numbers are scaled by 100. The easy test set contains only the same stickers seen during training, while the hard test set has unseen stickers. R@k is the recall rate of top-k predicted stickers and MRR the Mean Reciprocal Rank of ground truth stickers. The abbreviations \textit{ctx}, \textit{emo} and \textit{sem} correspond to the auxiliary task 1, 2 and 3 respectively in Section~\ref{method}.}
		\label{tab:sensitivity}}
\end{table*}

\section{Effect of Semantic Information}
\label{sec:ocr}

In our full model, we also added semantic labels to other tasks' inputs, i.e., the main task of context-sticker matching, the masked context prediction task and the sticker emotion classification task. It raises an interesting question of how the performance will change if we remove this information.
The result is shown in Table \ref{tab:result2}. As we can see, the sticker semantic prediction task is more beneficial for the easy test set, while adding OCR information to other tasks is more beneficial for the hard test set. We conjecture that because of the relatively small number of stickers (less than 300), it could be easier for the model to memorize the meaning of all stickers in the dataset, which potentially damages the model's generalization ability on unseen stickers in the hard test set. Adding OCR information for other tasks greatly alleviates this phenomenon because it could offer semantic labels for unseen stickers and enhance the model's generalization ability. 

We also tried to enhance the model's generalization ability by incorporating additional sticker-description pairs from another source into the sticker semantic prediction task. As Table \ref{tab:result2} shows, this method could increase the performance on the hard test set as expected, but the performance on the easy test set drops significantly, which may be attributed to the distribution difference between the additional data and our original data.

\section{Analysis of Sensitivity to the Text in Stickers}
\label{sec:sensitivity}
To explore whether stickers have text or not could affect the model's performance, we split each test set into two parts, i.e., stickers with recognized text labels versus those without text labels. The number of samples in each part is 2164 and 1051 for the easy set, and 4429 and 2599 for the hard set. We compare the performance of our full model with that of CLIP on the divided test sets in Table \ref{tab:sensitivity}. To avoid randomness in candidate set construction, we only compare the two models with the whole candidate set. In general, our full model and CLIP work better when the stickers have text, which suggests that the text in the sticker could help the model better understand the sticker.\footnote{Strictly speaking, there may be other factors that make the two parts inherently different in difficulty, but a fair comparison is difficult to make. Intuitively, stickers without text labels are generally harder to understand for the models.} However, our model is less sensitive to whether stickers have text or not according to the smaller difference value compared with CLIP. This implies our model is more robust to different kinds of candidate stickers.

\end{document}